\documentclass[letterpaper, 10 pt, journal, twoside]{IEEEtran}
\usepackage{amsmath,amsfonts}
\usepackage{algorithm}
\usepackage[caption=false,font=normalsize,labelfont=sf,textfont=sf]{subfig}
\usepackage{textcomp}
\usepackage{url}
\usepackage{verbatim}
\usepackage{graphicx}
\usepackage{paralist}
 
\usepackage{graphics} 
\usepackage{epsfig} 
\usepackage{mathptmx} 
\usepackage{times} 
\usepackage{amsmath} 
\usepackage{amssymb}  
\usepackage{cite}
\usepackage{algpseudocode}
\usepackage{balance}
\usepackage{multirow}
\usepackage{glossaries}
\usepackage{CJK}
\usepackage{booktabs}
\setacronymstyle{long-short}
\usepackage[dvipsnames]{xcolor}
\usepackage{hyperref}
\usepackage{enumerate}
\usepackage{microtype} 
\usepackage[export]{adjustbox}
\usepackage{dblfloatfix}
\DeclareUnicodeCharacter{2212}{-}
\usepackage{array}
\usepackage{hyperref}  
\usepackage{microtype} 

\begin{document}
 
\title{
A Reinforcement Learning Approach to Non-prehensile Manipulation through Sliding}
 
\author{Hamidreza Raei$^{1,2}$, Elena De Momi$^{2}$, Arash Ajoudani$^{1}$
\thanks{$^1$HRI$^2$ Lab, Istituto Italiano di Tecnologia, Genoa, Italy. {\tt\small hamidreza.raei@iit.it}}%
\thanks{$^{2}$ Department of Electronics, Information and Bioengineering, Politecnico di Milano, Milan, Italy.}
\thanks{This work was supported by the European Commission's Marie Skłodowska-Curie Actions (MSCA) Project RAICAM (GA 101072634) and Horizon Europe TORNADO (GA 101189557).}%
}
 


\maketitle
\begin{abstract}

Although robotic applications increasingly demand versatile and dynamic object handling, most existing techniques are predominantly focused on grasp-based manipulation, limiting their applicability in non-prehensile tasks. To address this need, this study introduces a Deep Deterministic Policy Gradient (DDPG) reinforcement learning framework for efficient non-prehensile manipulation, specifically for sliding an object on a surface. The algorithm generates a linear trajectory by precisely controlling the acceleration of a robotic arm rigidly coupled to the horizontal surface, enabling the relative manipulation of an object as it slides on top of the surface. Furthermore, two distinct algorithms have been developed to estimate the frictional forces dynamically during the sliding process. These algorithms provide online friction estimates after each action, which are fed back into the actor model as critical feedback after each action. This feedback mechanism enhances the policy's adaptability and robustness, ensuring more precise control of the platform's acceleration in response to varying surface condition. The proposed algorithm is validated through simulations and real-world experiments. Results demonstrate that the proposed framework effectively generalizes sliding manipulation across varying distances and, more importantly, adapts to different surfaces with diverse frictional properties. Notably, the trained model exhibits zero-shot sim-to-real transfer capabilities.

\end{abstract}

\section{INTRODUCTION}


For years, advancements in robotic manipulation have been concentrated on enhancing grasping dexterity and generalizing skills across a wide range of objects and tasks. These developments have significantly narrowed the gap between robotic manipulators and skilled humans, particularly in prehensile manipulation which are performed in slow and quasi-static operations. The ability to manipulate objects without grasping them is a decisive aspect of human dexterity. Despite the growing demand of involving robots in various applications where non-prehensile manipulation is the most effective way to perform the task, current algorithms developed for non-prehensile manipulation still fall short of achieving human-level performance. This limitation primarily stems from their dependence on oversimplified models and approximated physical parameters, which compromise the robustness of these methods when dealing with uncertainties, variations in object properties, and dynamic environmental conditions. Non-prehensile manipulation tasks are essential in various fields, such as logistics, cooking, and food delivery. Examples include sliding a pizza into an oven or flipping a burger in a pan. 

\begin{figure}[t]
\centering
    \includegraphics[trim=0.2cm 0.1cm 0.2cm 0.2cm, clip, width=0.42\textwidth, height=0.4\textheight]{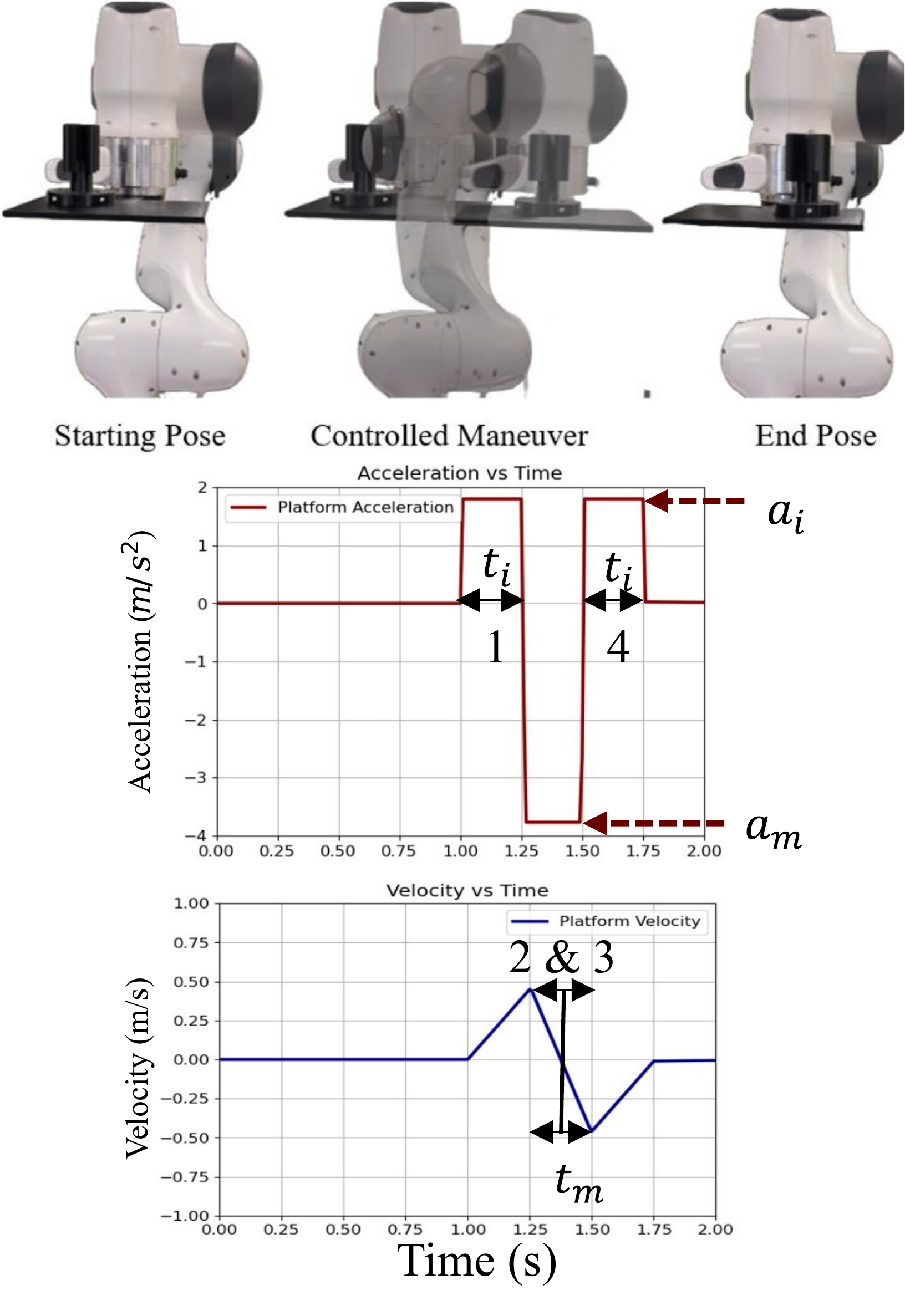}
    \vspace{-10pt}
    \caption{Illustration of the non-prehensile manipulation explored in this study: A robotic arm sliding an object using a controlled maneuver by following determined and commanding cartesian velocity to the robotic arm controller.}
    \label{fig:concept}

\end{figure}

To address these limitations in non-prehensile manipulation, this study investigates precise sliding control of objects on a horizontal surface using a robotic arm. This approach differs from other studies on non-prehensile manipulation methods that focus on preventing objects from sliding across a horizontal surface during the transportation \cite{c18}, as well as from previous work that uses external barriers to slide objects on the manipulator’s palm \cite{c1}. Our study aims to achieve accurate sliding displacements by controlling surface acceleration and deceleration (see Fig. \ref{fig:concept}), similar to moving a glass on a tray, without requiring prior knowledge of the precise friction coefficient and only by maneuvering the tray. To perform this, an actor-critic reinforcement learning (RL) framework is implemented in a simulation environment. The RL framework generates sequences of actions in a continuous action space to create precise linear trajectories, achieving the desired sliding displacement. Once trained, the actor model is transferred to real-world setup to be evaluated.

\textcolor{black}{
However, deploying policies trained in simulation directly into real-world scenarios often presents a significant challenge due to the sim-to-real gap. One approach to mitigating this gap is the use of higher-fidelity simulation environments, which more accurately capture subtle real-world conditions \cite{c6}. In this study, we employ MuJoCo as our simulation environment due to its high physical accuracy. In robotic manipulation complex physical variables, such as surface roughness, and contact mechanics, are often challenging to measure accurately and can affect the performance of the algorithm's real world implementation. Yet, these factors are often approximated or overlooked altogether \cite{c2}. In non-prehensile manipulation, the dynamic friction between surfaces is particularly critical, in contrast to manipulation through grasping where static friction plays the key role. Most simulation environments utilize the coulomb friction model which is known for its inaccuracy in dynamic friction. Another common solution to bridge the gap between simulation and reality is domain randomization, which randomizes the dynamics of the environment and exposes the RL framework to a diverse set of environments in the training phase \cite{c5,c6}.}

In addition to measures taken to reduce the sim-to-real gap in the face of physical uncertainties, we further enhance robustness by estimating and incorporating real-time friction feedback into the control algorithm. To achieve this, we introduce two friction inference methods: an analytical approach and a data-driven Long Short-Term Memory (LSTM) model. In summary, the contributions of this work can be listed as follows:


\begin{itemize}
    \item Training an actor-critic RL model which is robust to surface friction and can slide an object on a surface by generating linear trajectories.

    \item Training an LSTM model to infer surface friction from time-series kinematic data, to infer the friction between the two surfaces.

    \item Developing an analytical friction inference approach using kinematic data to robustly estimate friction without learned models.

    \item Assessing the effect of domain randomization and friction inference methods on the performance of zero-shot sim-to-real transfer.
\end{itemize}

\section{Related Works}

\subsubsection{\textbf{Non-prehensile manipulation}} Non-prehensile manipulation, or the ability to manipulate objects without grasping, is crucial for tasks involving hard-to-grasp objects or cluttered workspaces \cite{c7, c8}. This types of manipulation includes throwing \cite{c21}, pushing \cite{c20}, rolling \cite{c19}, and sliding \cite{c1} among others. Several studies have explored non-prehensile manipulation of an object on horizontal surfaces \cite{c1, c18, c9}. One of these studies \cite{c18} primarily focused on transporting an object on a horizontal surface while ensuring no sliding or slippage occurs during the task. In contrast, the objective of our work deliberately leverages sliding as the primary method for moving the object across the surface. The other study \cite{c1}, investigates the control approach of sliding an object on the surface and impressive controllability and performance were demonstrated. It is important to note that this study assumes the presence of external barriers to assist in performing the sliding task and incorporates precise knowledge of the surface friction coefficient (\textit{$\mu$}) into its model-based controller. While this setup enables impressive controllability and performance, it inherently limits the approach's ability to generalize to scenarios without external constraints or surfaces with unknown friction properties. Another study in this area \cite{c9}, avoided external barriers but lacked real-time friction adaptation, which our work addresses through on-line friction estimation and robust RL-based control.


\subsubsection{\textbf{RL for continuous action space}} Reinforcement learning algorithms have long been studied and applied in simplified environments, such as the Atari games discussed in \cite{c10}. In these settings, they have achieved notable success, often surpassing human experts in performance. However, these environments rely on discrete action spaces, and for most control problems in real world applications, continuous action space is necessary. Several methods have been introduced to extend reinforcement learning to continuous action spaces. Among the most prominent are actor-critic models like Deep Deterministic Policy Gradient (DDPG) \cite{c11} and Proximal Policy Optimization (PPO) \cite{c12}, both of which are proven highly effective for handling continuous control tasks. Thus, these method have been utilized more frequently in recent studies in robotics such as training an actor-critic model for non-prehensile manipulation where the end-effector can interact with the object on an external surface, to perform the manipulation, in which they achieved a 50\% success in zero-shot sim-to-real transfer over unseen objects \cite{c7}.
Actor-critic RL models can be applied to other applications as well, such as robotic motion planning in dynamic environments \cite{c14} and drones learning to fly through gates autonomously \cite{c13}.

\section{Methodology}

In this study, we propose a discrete-time control strategy for sliding manipulation, inspired by the way humans perform rapid, step-wise adjustments to slide objects on a surface. Our approach leverages an iterative process that adapts to surface friction by incorporating feedback from past actions and real-time observations, particularly when surface friction estimation is inaccurate. Following this, we present the reinforcement learning framework, providing details about the simulation environment and the architecture of the actor and critic models. Finally, we introduce two distinct online algorithms for friction estimation: an analytical approach and a data-driven LSTM approach, both designed to infer friction based on the action given to the manipulator and the kinematic data of relative motion between the object and the surface.

\begin{figure}
\centering
    \includegraphics[trim=0.2cm 0.2cm 0.2cm 0.2cm, clip, width=0.16\textwidth, height=0.1\textheight]{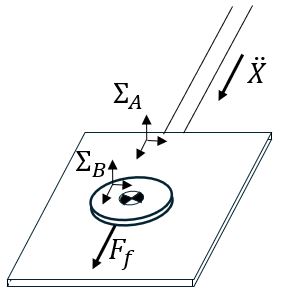}
    \caption{A simple Schematic of the platform and the sliding object where \textit{$\Sigma_B$} is the coordinate system fixed on the sliding object and  \textit{$\Sigma_A$} is the coordinate system fixed on the platform.}
    \label{fig:Schematic}
\end{figure}

\subsection{\textbf{Problem Statement and Assumptions} \label{sec:assumptions}} 

Initially, we define a strategy under which the sliding manipulation on a surface by linear maneuver can be mathematically formulated. In Fig. \ref{fig:Schematic} the parameters used in the formulation for sliding are illustrated. The condition for the object to start sliding on the surface, is the required inertial force being more than the maximum static frictional force, shown in (\ref{eq:conditional}), and once that is satisfied, the relative equation of motion for the object with respect to the surface is illustrated in (\ref{eq:motion}).
\begin{equation}
    F_{f_{max}} = \mu_s g \hspace{1cm} |\mathbf{{\ddot{X}}}| > \mu_s g,
    \label{eq:conditional}
\end{equation}

\begin{equation}
    {\ddot{x}}_A^B = -\mathbf{\ddot{X}} - \frac{\dot{x}_A^B}{|\dot{x}_A^B|} \mu_k g.
    \label{eq:motion}
\end{equation}

To simplify the problem of the sliding object, assumptions are made regarding the input command \textit{$\mathbf{\ddot{X}}$} which defines the linear acceleration of the supporting surface. The surface starts from a stationary position, accelerates, decelerates, and returns to its initial resting pose. This motion defines a step in the DDPG reinforcement learning framework. The relative displacement of object B with regards to the surface A is defined in (\ref{eq:displacement}),


\begin{equation}
x_A^B =
\begin{cases} 
0 &  |\ddot{X}| \leq \mu_s g \quad  \\ 
-\iint \ddot{X} \, dt - \iint \frac{|\dot{x}_A^B|}{\dot{x}_A^B} \mu_k\,g \, dt 
&  |\ddot{X}| > \mu_s g \quad 
\end{cases}.
\label{eq:displacement}
\end{equation}

\subsection{\textbf{Generated Action into Linear Trajectory} \label{sec:action_to_trajectory}} 


Fig. \ref{fig:concept} shows the determined command for a desired trajectory, and the hardware setup used to implement the experiment. The action generated by the RL agent comprises three parameters which are two accelerations and one time duration defined in Fig. \ref{fig:concept} denoted as (\textit{$a_i, a_m, t_m$}). To elaborate the maneuver using these parameters, we need to write down the velocity of the surface as a function of time derived from this command, which is shown in \eqref{eq:velocity_time}. The maneuver is divided into four phases. The transition from one phase to another happens when there is a change of commanded acceleration or change of direction of the motion of the platform, as illustrated in the platform velocity versus time in Fig. \ref{fig:concept}. At \textit{$t = t_i + \frac{t_m}{2}$} the platform change its direction of motion and that means \textit{$\dot{x}_A(t) = 0$}.

By solving (\ref{eq:velocity_time}), for \textit{$t = t_i + \frac{t_m}{2}$}, the undetermined parameter \textit{$t_i$} in the action sequence can be explicitly calculated.  The resulting expression for \textit{$t_i$} is provided in (\ref{eq:time_d}). The end of the second phase marks the maximum distance from the starting pose which is crucial as the robotic arm has a limited range of motion. The required range of motion based on the generated action is obtained in (\ref{eq:distance}). This formulation presented for \textit{$t_i$} ensures that the robot begins and concludes the maneuver at the same pose, adhering to the initial assumption discussed in \ref{sec:assumptions}. The calculated \textit{$\dot{x_A}$} will be commanded to the franka cartesian velocity controller to be executed.

\begin{equation}
\scalebox{0.92}{$
\dot{x}_A(t)=
\begin{cases}
a_i\,t & 0 \leq t \leq t_i \\[3pt]

a_i\,t_i
+ {a_m}(t - t_i) & t_i \leq t \leq t_i+\frac{t_m}{2}\\[3pt]

a_i\,t_i
+ {a_m} \frac{t_m}{2} + {a_m}(t - t_i - \frac{t}{2}) & t_i+\frac{t_m}{2} \leq t \leq t_i + t_m\\[3pt]

a_i\,t_i
+ {a_m}\,t_m + a_i\, t & t_i+t_m \leq t \leq 2 t_i + t_m\\[3pt]
\end{cases}
$}
\label{eq:velocity_time}
\end{equation}

\begin{equation}
    t_i = \frac{t_m}{2} \left| \frac{a_m}{a_i} \right|,
    \label{eq:time_d}
\end{equation}

\begin{equation}
    \int_{0}^{t_i + \frac{t_{_m}}{2}} \dot{x}_A(t) \, dt = \frac{a_i {t_i}^2}{2} + \frac{a_m\, t_m^2}{8}.
    \label{eq:distance}
\end{equation}

\subsection{\textbf{Reinforcement Learning Framework}}

To handle the system's actions in a continuous space, we utilize the DDPG framework, an actor-critic RL approach. For training, the MuJoco simulator \cite{c3} is selected, due to its high-fidelity physics simulation, and light-weight computational load.  The DDPG model is an off-policy, deterministic policy gradient algorithm \cite{c15}, that generates the same action for a given state during evaluation, making it more predictable than stochastic policy models. The deterministic nature of DDPG can aid sim-to-real deployment when combined with techniques like domain randomization and robust policy training, as demonstrated in \cite{c22, c23}. In this method, the objective function is defined as (\ref{eq:objective}). The actor network, represented as \textit{$\pi(s|\theta^\pi)$}, is updated by applying the gradient of the objective function \textit{$J(\theta^{\pi})$} with regards to its actor network parameters, denoted as \textit{$\theta^\pi$} shown in (\ref{eq:gradient}). This update involves two components: the gradient of the critic \textit{$Q(s, a|\theta^Q)$} with respect to the action \textit{$a$}, evaluated at the policy output \textit{$a = \pi(s|\theta^\pi)$}, and the gradient of the policy’s output \textit{$\pi(s|\theta^\pi)$} with respect to its parameters \textit{$\theta^\pi$}. The critic \textit{$Q(s, a|\theta^Q)$} learns to approximate the expected cumulative reward using the Bellman equation (\ref{eq:bellman}), which defines the relationship between immediate and future rewards. Together, these components ensure the policy is improved to maximize the expected return.
 

\begin{equation}
    J(\theta^\pi) = \mathbb{E}_{s \sim \mathcal{D}} \left[ Q(s, \pi(s|\theta^\pi) |\theta^Q) \right],
    \label{eq:objective}
\end{equation}

\begin{equation}
    \nabla_{\theta^\pi} J(\theta^\pi) = \mathbb{E}_{s \sim \mathcal{D}} \left[ \nabla_a Q(s, a | \theta^Q) \nabla_{\theta^\pi} \pi(s | \theta^\pi) \right],
    \label{eq:gradient}
\end{equation}

\begin{equation}
    Q(s, a) = r(s, a) + \gamma \mathbb{E}_{s'} \left[ Q(s', \pi(s')) \right].
    \label{eq:bellman}
\end{equation}

It is worth mentioning that this framework is designed for zero-shot sim-to-real transfer of the learned policy. Below, we detail the components of this RL framework.

\subsubsection{\textbf{Data structure}}
We are using a symmetric implementation of the actor-critic model, where both have access to the same set of state from the environment. The state for this structure includes, desired remaining distance for displacement denoted as \textit{$D_{des}$}, up to three previous actions, (\textit{$A_{i-1}, A_{i-2}, A_{i-3}$}) and the relative displacement caused by each action (\textit{$D_{i-1}, D_{i-2}, D_{i-3}$}), and a guesstimate of the friction coefficient shown as \textit{$\mu_e$}. In simulation, all states are accurately available, whereas in real-world implementation, relative position is tracked via a motion capture system, and friction is estimated through extensive experimentation.

\subsubsection{\textbf{Action space}}

At each step, the learned policy generates a three-dimensional action that defines the trajectory using three parameters \textit{$a_i$} (initial acceleration), \textit{$a_m$} (maximum acceleration), and \textit{$t_m$} (time duration). This action turns into a linear trajectory as stated in \ref{sec:action_to_trajectory}. The generated action is constrained by acceleration and motion range limits to comply with the robotic arm's physical capabilities, as stated in \eqref{eq:action_space},
\begin{equation}
\text{Action Space} = 
\left\{
\begin{aligned}
&|a_i|, |a_m| \leq 4.2 \, \text{m/s}^2 \\
&t_m < 2.0 \, \text{s} \\
&a_i\, a_m \leq 0
\end{aligned}.
\right.
\label{eq:action_space}
\end{equation}

\begin{figure}
\centering
    \includegraphics[trim=0.0cm 0.0cm 0.0cm 0.0cm, clip, width=0.45\textwidth, height=0.22\textheight]{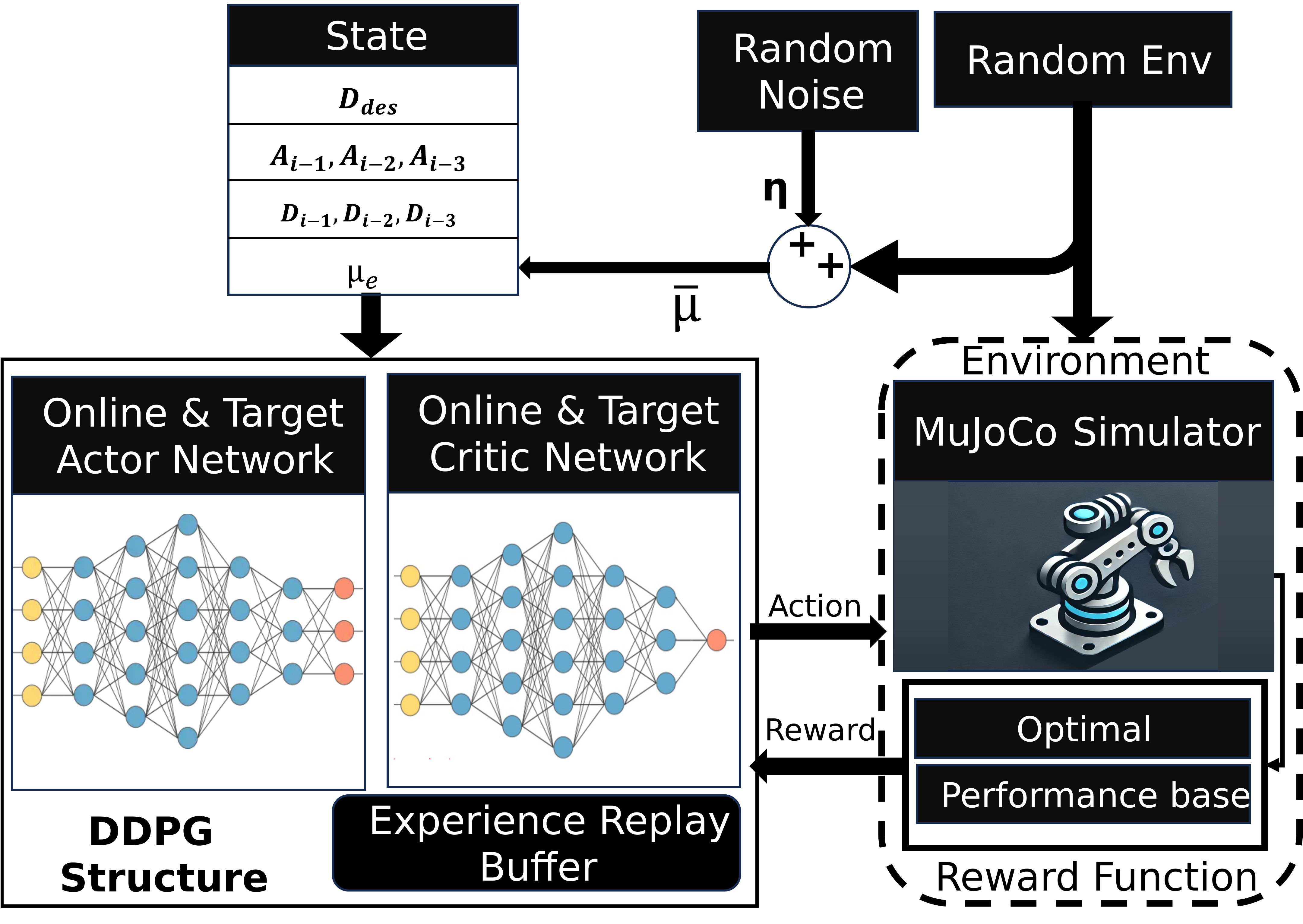}
    \caption{This figure illustrate, the states, the environment random initialization and domain randomization on friction parameter, in the DDPG training framework.}
    \label{fig:training_framework}
\end{figure}

\subsubsection{\textbf{Reward function and training}}

By formulating the sliding task, assuming the accurate knowledge of the friction coefficient in Coulomb model, the optimal solution can be acquired. Although this optimal solution works perfect in simulation, it cannot satisfy the task in the real world implementation considering the prior assumptions. As a result a progressive \textit{\textbf{reward shaping}} approach \cite{c17} is implemented, and the training is divided into the several steps. First step is mimicking the analytical solution for the Coulomb friction model for a single distance that is reinforced by shaping the reward function to incentivize proximity of the action parameters to optimal values which are (\textit{$a_i = \mu_s\,g $}, \textit{$a_m = 4.2 \frac{m}{s^2}$}) and \textit{$t_m$} can be obtained from the expansion of \eqref{eq:displacement} which is illustrated in \eqref{eq:relative_vel} for the relative velocity and \eqref{eq:relative_motion} for the relative displacement from the start to the end of each phase for all of the four phases stated in \eqref{eq:velocity_time}. The resulting displacement is summation of relative displacement of all the four phases. If any of the acceleration is lower than the \textit{$ \mu_s\,g$}, the corresponding displacement of that phase is zero. The complexity of this maneuver is reflected in \eqref{eq:relative_acceleration}, where the object's acceleration varies within the same phase, even without changes in the surface's motion direction or acceleration.





\begin{equation}
    \scalebox{0.8}{$
\ddot{x}^{B}_A(t)=
\begin{cases}
{a_i - \mu_k\,g}&  \, 0 \leq t \leq t_i \\[3pt]

{a_m -  \, \frac{\left| \dot{x}^B_A(t) \right| \mu_k\,g}{\dot{x}^B_A(t)}} & t_i < t \leq t_i + \frac{t_m}{2} \\[3pt]

{a_m -  \, \frac{\left| \dot{x}^B_A(t) \right| \mu_k\,g}{\dot{x}^B_A(t)}} & t_i + \frac{t_m}{2} < t \leq t_i + t_m \\[3pt]

{a_i - \frac{\left| \dot{x}^B_A(t) \right| \mu_k\,g}{\dot{x}^B_A(t)}} & t_i + t_m < t \leq 2t_i + t_m\\[3pt]
\end{cases},
$}
\label{eq:relative_acceleration}
\end{equation}

\begin{equation}
\scalebox{0.8}{$
\dot{x}^{B}_A(t)=
\begin{cases}
({a_i - \mu_k\,g})\,{t} & 0 \leq t \leq t_i \\[3pt]

\dot{x}^{B}_A(t_i) + \,(\ddot{x}^{B}_A(t))\,({t - t_i}) & t_i < t \leq t_i + \frac{t_m}{2} \\[3pt]

\dot{x}^{B}_A(t_i + \frac{t_m}{2}) + (\ddot{x}^{B}_A(t)))\,(t - t_i +\frac{t_m}{2}) & t_i + \frac{t_m}{2} < t \leq t_i + t_m\\[3pt]

\dot{x}^{B}_A(t_i + t_m) + (\ddot{x}^{B}_A(t))\,(t-t_i-t_m) & t_i + t_m < t \leq 2t_i + t_m \\[3pt]
\end{cases},
$}
\label{eq:relative_vel}
\end{equation}

\begin{equation}
\scalebox{0.8}{$
\Delta \,{x}^{B}_A=
\begin{cases}
\frac{(\ddot{x}^{B}_A(t))}{2}\,{t_i}^2 & \left |a_i\right | > \mu_s\,g \\[3pt]

{\dot{x}^{B}_A(t_i)}\,\frac{t_m}{2}
+ (\ddot{x}^{B}_A(t))\frac{t_m^2}{8} & \left |a_m\right | > \mu_s\,g \\[3pt]

{\dot{x}^{B}_A(t_i+\frac{t_m}{2})}\,\frac{t_m}{2}
+ (\ddot{x}^{B}_A(t))\frac{t_m^2}{8} & \left |a_m\right | > \mu_s\,g \\[3pt]

{\dot{x}^{B}_A(t_i+t_m)}\,t_i
+ \frac{(\ddot{x}^{B}_A(t))}{2}\,{t_i}^2 & \left |a_m\right | > \mu_s\,g\\[3pt]
\end{cases}.
$}
\label{eq:relative_motion}
\end{equation}

As the performance of the actor model increases the reward function automatically switches to a performance based function, which takes into account the accuracy of displacement through sliding, maximum range-of-motion defined in \eqref{eq:distance}, the required time and number of steps for a desired displacement. In the second step, we give various distances for each episode ranging from 0.02 \textit{$m$} to 0.2 \textit{$m$}, and the reward function remains performance base. In the third step, not only various distances but also various friction coefficient (\textit{$\mu$}) is given to the simulation environment for each episode ranging from 0.05 to 0.45.

\begin{figure*}[!htbp]
    \centering
    \includegraphics[trim=0 0.0cm 0 0.0cm, clip, width=0.9\textwidth, height=0.34\textheight]{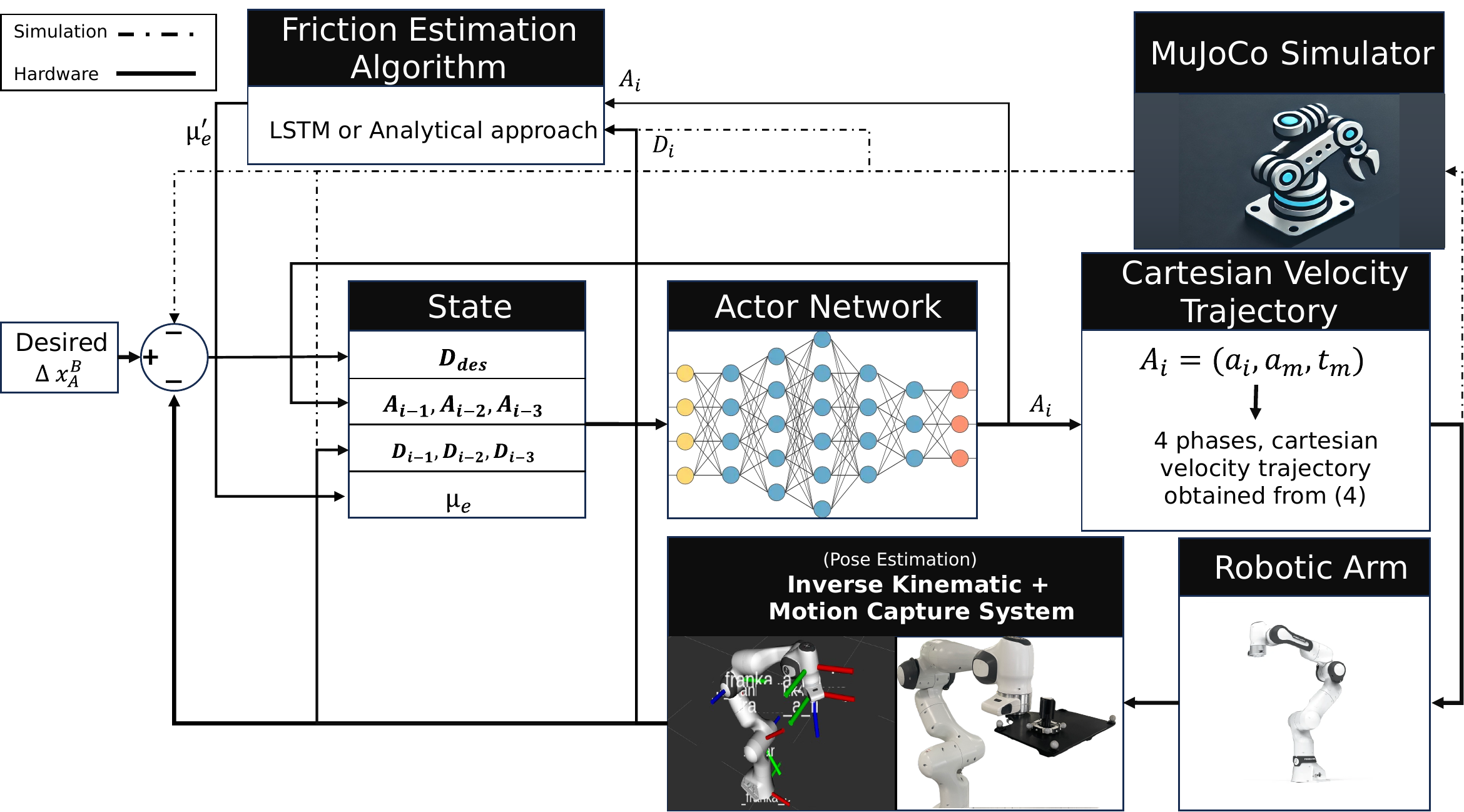}
    \caption{A schematic of the evaluation framework. Connections between components for real-world experiments (solid lines) and simulations (dashed lines) are shown. The key difference lies in pose estimation: real-world setups rely on motion capture systems and inverse kinematics to determine the pose of the end-effector and the object, while simulations provide these parameters directly.}

    \label{fig:block_diagram_evaluation}
\end{figure*}

\subsubsection{\textbf{Domain Randomization}}

Upon completing the training phase, where the model successfully generalizes its actions across varying distances and friction coefficients (\textit{$\mu$}), domain randomization is introduced to enhance robustness and sim-to-real transferability \cite{c6}. For this problem, friction is identified as the most critical parameter influencing the sliding task. By introducing variability in \textit{$\mu$} the goal is to enable the model to handle a wide range of frictional conditions and prepare it for deployment in real-world environments, where the Coulomb friction model may not fully capture the underlying dynamics. To achieve this, we added a noise according to \eqref{eq:noise} to the friction coefficient, provided to the state for the actor and the critic model in two stages. In the first stage, \textit{$\eta \in [0.05, 0.1]$} and in the second stage \textit{$\eta \in [0.05, 0.15]$}. As highlighted in \cite{c24}, excessive randomization can lead to overly conservative policies, which is why the range was incrementally adjusted. In Fig. \ref{fig:training_framework}, the training structure for the DDPG RL framework is depicted.

\begin{equation}
    \Bar{\mu} = \mu \pm \eta
    \label{eq:noise}
\end{equation}

\subsection{\textbf{Friction Inference Algorithms}
\label{sec:algorithms}} 

This study introduces two algorithms that estimate friction using kinematic data of the relative motion and the actor model's commands, providing feedback on surface roughness to enhance the actor model's performance.

\subsubsection{\textbf{Analytical approach}} 
In this approach, the relative equations of motion stated in \eqref{eq:relative_vel} and \eqref{eq:relative_motion} are utilized to solve \textit{$\mu_k$} for the first three phases of the maneuver, given the known \textit{$x^B_A$} from the simulation data or motion capture system in real world experiment. With the generated action provided, the only unknown parameter is the friction coefficient, \textit{$\mu_k$}, as detailed in Algorithm \ref{alg:friction_estimation}.
\begin{algorithm}
\caption{Analytical Friction Estimation}
\label{alg:friction_estimation}
\textbf{Input:} Relative displacement \(x^B_A(t)\) (measured via OptiTrack or simulation), generated action (\(a_i, a_m, t_i, t_m\)) \\
\textbf{Output:} Friction coefficient \(\mu_e\)
\begin{algorithmic}[1]
\If{\(x^B_A(t = t_i) > 1.0 \, \text{cm}\)}
    \State \textit{\textbf{$\mu_e \leftarrow$ Solve }}
    [$\frac{1}{2}\,(a_i - \mu_k \, g)\, t^2 = x^B_A(t_i)$] for \textit{$\mu_k$}

\ElsIf{\(x^B_A(t_i + t_m) - x^B_A(t_i) > 1.0 \, \text{cm}\)}
    \State \textit{\textbf{$\mu_e \leftarrow$ Solve}}
    [$\dot{x}^B_A(t_i)\,t_m + \frac{(a_m - \mu_k \, g )\, t_m^2}{2}= x^B_A(t_i + t_m) - x^B_A(t_i)$] for \textit{$\mu_k$}
    
\Else
    \State  
    $\mu_e \leftarrow 1.1 \, \frac{a_m}{g}$
\EndIf
\end{algorithmic}
\end{algorithm}

\subsubsection{\textbf{Data-driven LSTM approach}} 

In this approach, the training data for the LSTM were exclusively generated from the simulation environment. The input to the LSTM consists of a time series of commanded accelerations applied to the surface and the relative velocity between the object and the surface for a duration of 2.0\textit{s}. The target output is the friction coefficient corresponding to the simulation environment. The network consists of four recurrent layers with 1024, 512, 256, and 128 neurons, each utilizing kernels, recurrent kernels, and biases to process inputs and capture temporal dependencies. Fully connected layers follow, progressively reducing dimensionality, with the final output layer predicting a single friction coefficient value.

\subsection{\textbf{Evaluation Setup}}

The DDPG framework is evaluated in both simulation and real-world experiments, as shown in Fig. \ref{fig:block_diagram_evaluation}. Although in the simulation setup, all required parameters can be obtained through the simulation itself, for the real-world implementation, parameters regarding motion of the surface is obtained through inverse kinematics of the franka arm and the parameters regarding the relative motion of the object on the surface is obtained through motion capture system. The evaluation setup is capable of including or excluding the friction estimation algorithms, and its effect is examined in the following section.


\section{Experiment and Results}

The experiment in this study consists of two main phases to comprehensively evaluate the proposed framework. In the first phase, simulations are conducted to assess the actor network's ability to generalize the sliding task across varying environmental parameters, such as displacement distances, surface friction, and inaccuracies in estimated friction coefficients. This simulation phase benchmarks the RL model's performance against analytical solutions, highlighting its adaptability under uncertainty in conditions. Additionally, it evaluates the friction estimation algorithms, and its effect on overall proposed framework. The second phase transitions the trained model into real-world scenarios to evaluate its robustness and zero-shot sim-to-real transfer capabilities. This phase involves testing the model on hardware as shown in Fig. \ref{fig:concept} with varying surface properties, validating its capacity to replicate desired behaviors without additional fine-tuning.

\subsection{ Simulation Evaluation}
\label{sec:Sim_result}

\begin{figure}
\centering
    \includegraphics[trim=0.05cm 0.05cm 0.05cm 0.05cm, clip, width=0.46\textwidth, height=0.3\textheight]{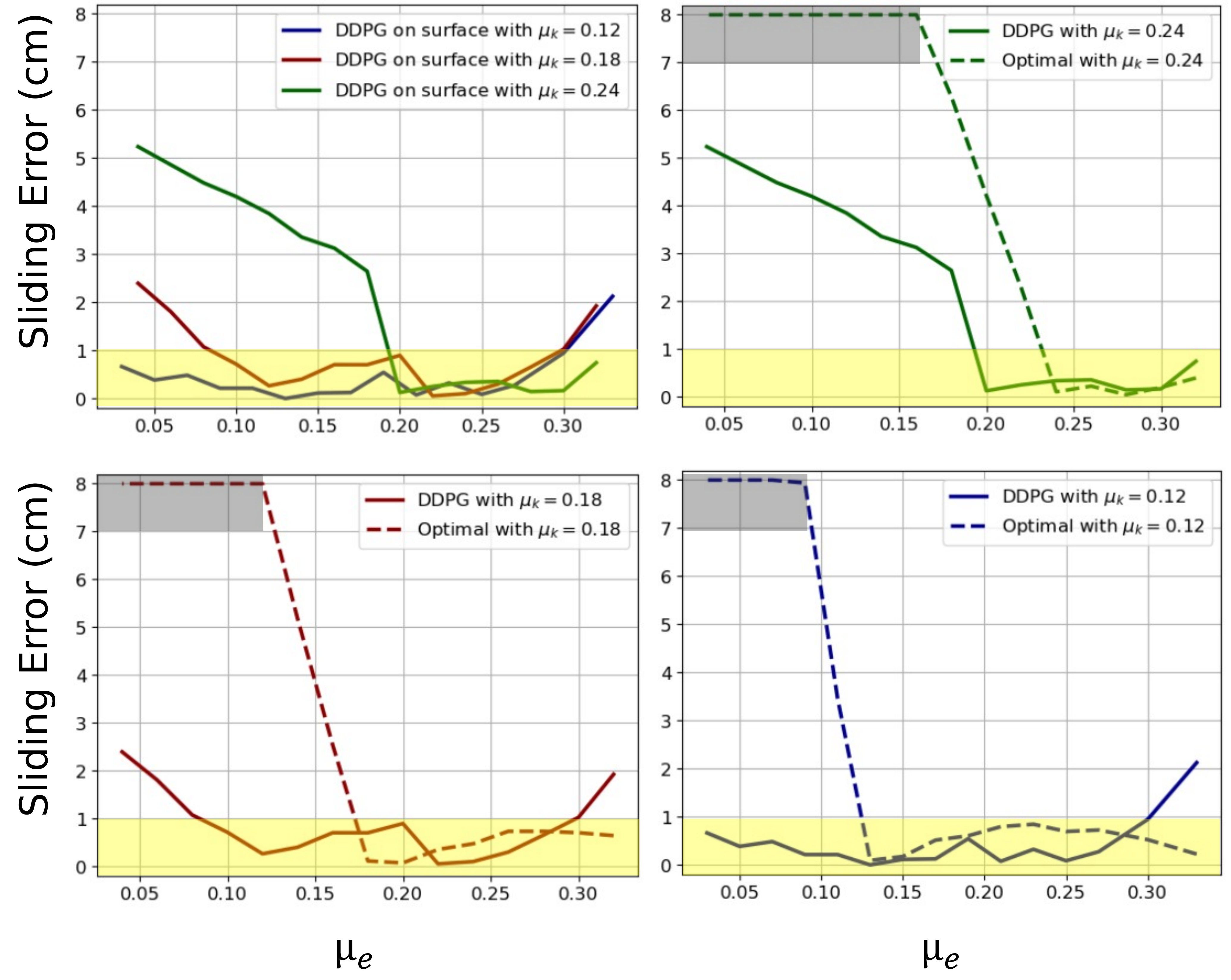}
    \caption{This plots illustrate performance of the actor network in sliding displacement under friction coefficient errors, compared to the analytically calculated optimal solution. Yellow highlights show the acceptable margin for errors, while gray highlights denote complete action failure to move the object.}
    \label{fig:DDPG_vs_optimal}
\end{figure}

This experiment aims to evaluate how effectively the actor network performs a sliding task over a fixed distance when exposed to varying friction coefficients. The friction coefficients, provided as inputs to the network’s state (shown in Fig. \ref{fig:block_diagram_evaluation}). The evaluation focuses on analyzing the error in the initial action generated by the actor network and its dependency on the provided friction coefficient. Additionally, it examines the network's ability to generalize its performance across diverse friction conditions. 

To benchmark its performance, the DDPG model is compared against an analytical approach that provides optimal control for the sliding task, as defined by \eqref{eq:relative_motion}. The experiment is conducted across three surfaces with distinct friction coefficients (\textit{$\mu_k$}), where the friction values provided to the network's state range between \textit{$\mu_e$} = (0.04, 0.32). Fig. \ref{fig:DDPG_vs_optimal} highlights the comparative performance of the DDPG model and the analytical approach. The actor trained within the DDPG framework demonstrates robust performance, showing significantly reduced sensitivity to variations in friction coefficients compared to the analytical optimal solution. Additionally, the results indicate that across all three surfaces, for \textit{$D_{des} = 8\,cm$} the actor network is capable of performing the task completely regardless of the error between \textit{$\mu_k$} and \textit{$\mu_e$}, provided the discrepancy remains within a range of \textit{$|\mu_k - \mu_e| \leq 0.05$}. Notably, the actor model avoids "dead zones"—situations where the generated action fails to slide completely—even under challenging conditions. 

\begin{figure}
\centering
    \includegraphics[trim=0.01cm 0.01cm 0.01cm 0.15cm, clip, width=0.48\textwidth, height=0.13\textheight]{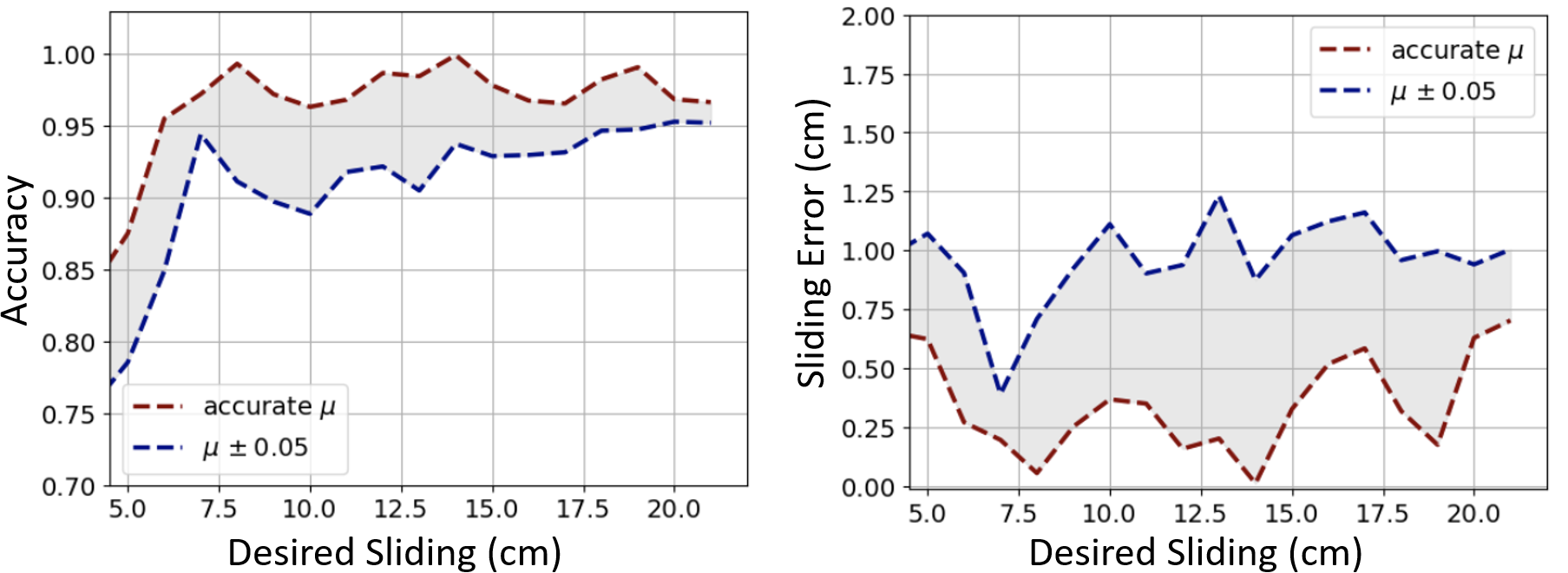}
    \caption{The performance of the actor model across different desired sliding distances (\textit{$D_{des}$}), evaluated under conditions of accurate and perturbed friction coefficient estimation.}
    \label{fig:DDPG_various_distances}
\end{figure}

\begin{figure}
\centering
    \includegraphics[trim=0.01cm 0.01cm 0.01cm 0.15cm, clip, width=0.48\textwidth, height=0.13\textheight]{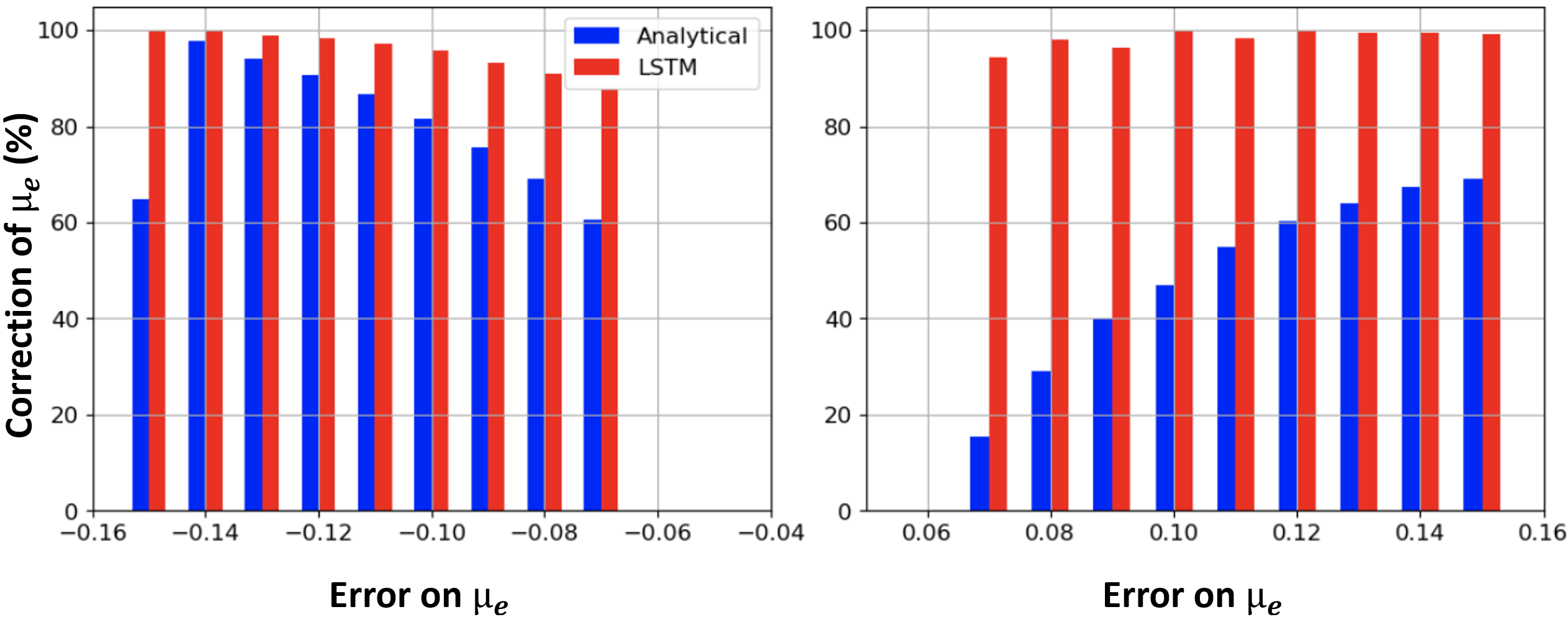}
    \caption{This plot illustrates how each friction estimation algorithm reduces the error in the estimated friction coefficient (\textit{$\mu_e$}), across a range of errors \textit{$|\mu_e - \mu_k|$} $\in  [0.07, 0.15]$. The $y$ axis  shows the percentage improvement in the accuracy of the estimated friction coefficient for each model (LSTM and Analytical), where higher values indicate better performance.}
    \label{fig:friction_estimate_acc}
\end{figure}

The second experiment evaluates the model's ability to generalize across various desired displacements (\textit{$D_{des}$}) in a simulation environment with a friction coefficient of \textit{$\mu_k = 0.24$}. As shown in Fig. \ref{fig:DDPG_various_distances}, the results highlight the model's accuracy (\textit{$1 -\frac{\left | \Delta x^B_A - D_{des} \right |}{|D_{des}|}$}) and the absolute error (\textit{$\left | \Delta x^B_A - D_{des} \right |$}) for both accurate (\textit{$\mu_e = \mu_k$}) and slightly deviated (\textit{$\mu_e = \mu_k \pm 0.05$}) friction estimates. The model achieves high precision, with errors of \textit{$\left | \Delta x^B_A - D_{des} \right | \leq 1 cm$} for accurate \textit{$\mu_e$} and \textit{$\left | \Delta x^B_A - D_{des} \right | \leq 1.25 cm$} for slightly deviated \textit{$\mu_e$}. It is important to note that all results presented in Fig. \ref{fig:DDPG_vs_optimal} and Fig. \ref{fig:DDPG_various_distances} focus exclusively on evaluating the performance of the actor network during its initial step.

\begin{figure}
\centering
    \includegraphics[trim=0.01cm 0.01cm 0.01cm 0.15cm, clip, width=0.48\textwidth, height=0.16\textheight]{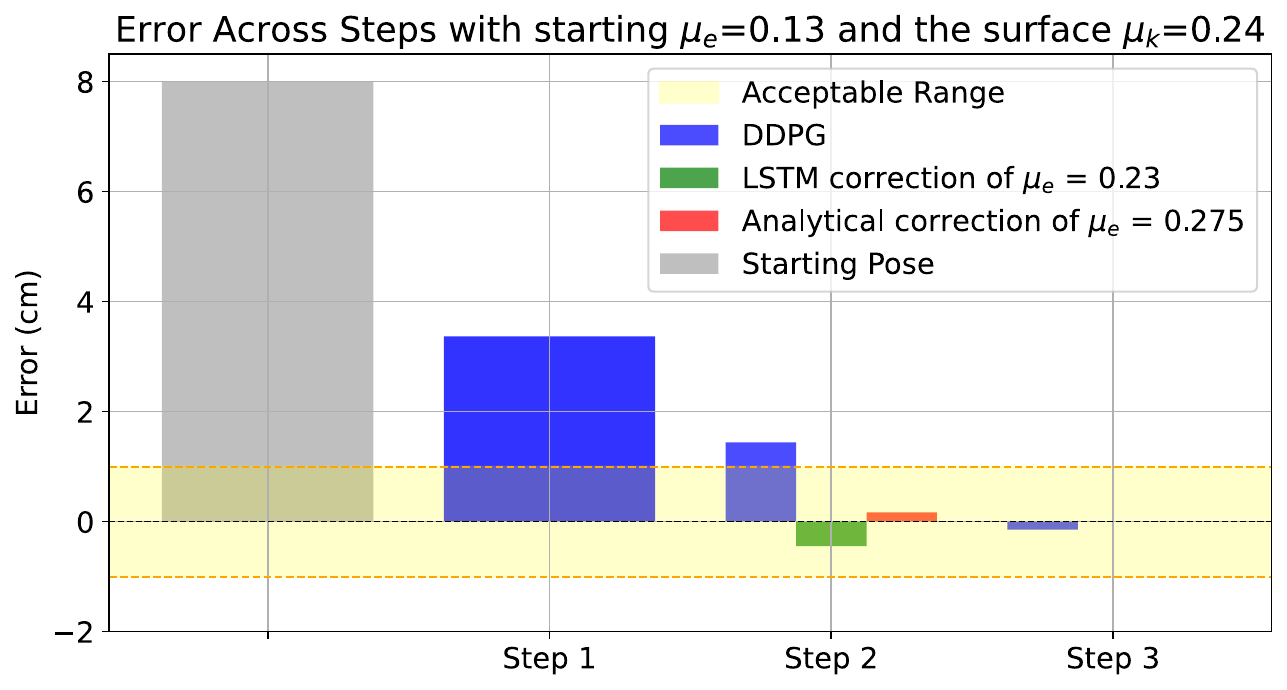}
    \caption{Comparison of the impact of friction estimation algorithms (analytical and LSTM) on task performance when the actor network starts with significant friction estimation error in simulation experiments.}
    \label{fig:sim_estimation_effect}
\end{figure}

Finally, the friction estimation algorithms are assessed for their accuracy and their influence on the overall performance of the evaluation framework. The assessment begins with the DDPG framework executing a sliding task with a desired displacement of \textit{$D_{des} = 8\,cm$} in an environment characterized by \textit{$\mu_k = 0.24$}. Following this, kinematic data, including the relative displacement (\textit{$x_A^B$}), relative velocity (\textit{$\dot{x}_A^B$}), and platform acceleration (\textit{$\ddot{x}_A$}), are utilized as per the methods described in \ref{sec:algorithms}. These data are applied to minimize the error between actual and initial estimation for friction coefficient denoted as (\textit{$\mu_k$}) and (\textit{$\mu_e$}) respectively. Assuming \textit{$\mu_e'$} shows output of friction estimation algorithms after the first step, the accuracy of correcting friction coefficient for each algorithm is calculated by \eqref{eq:correction percentage} and is presented in Fig. \ref{fig:friction_estimate_acc}, demonstrating that the data-driven LSTM approach outperforms the analytical method in the simulation setup. Furthermore, Fig. \ref{fig:sim_estimation_effect} shows a scenario where \textit{$\mu_k = 0.24$} and \textit{$\mu_e = 0.13$} is given as starting state. It demonstrates how the improved friction estimations enhance the accuracy of sliding displacement in subsequent steps compared to the standalone performance of the actor network. The results reveal that friction estimation not only reduces the error in the following step but also enables the task to be completed in less number of steps compared to the standalone actor network.

\begin{equation}
    \text{Correction of \textit{$\mu_e$}} = (1 - \frac{|\mu_k - \mu_e'|}{|\mu_k - \mu_e|}) 
    \label{eq:correction percentage}
\end{equation}



\begin{figure}
\centering
    \includegraphics[trim=0.01cm 0.01cm 0.01cm 0.15cm, clip, width=0.51\textwidth, height=0.32\textheight]{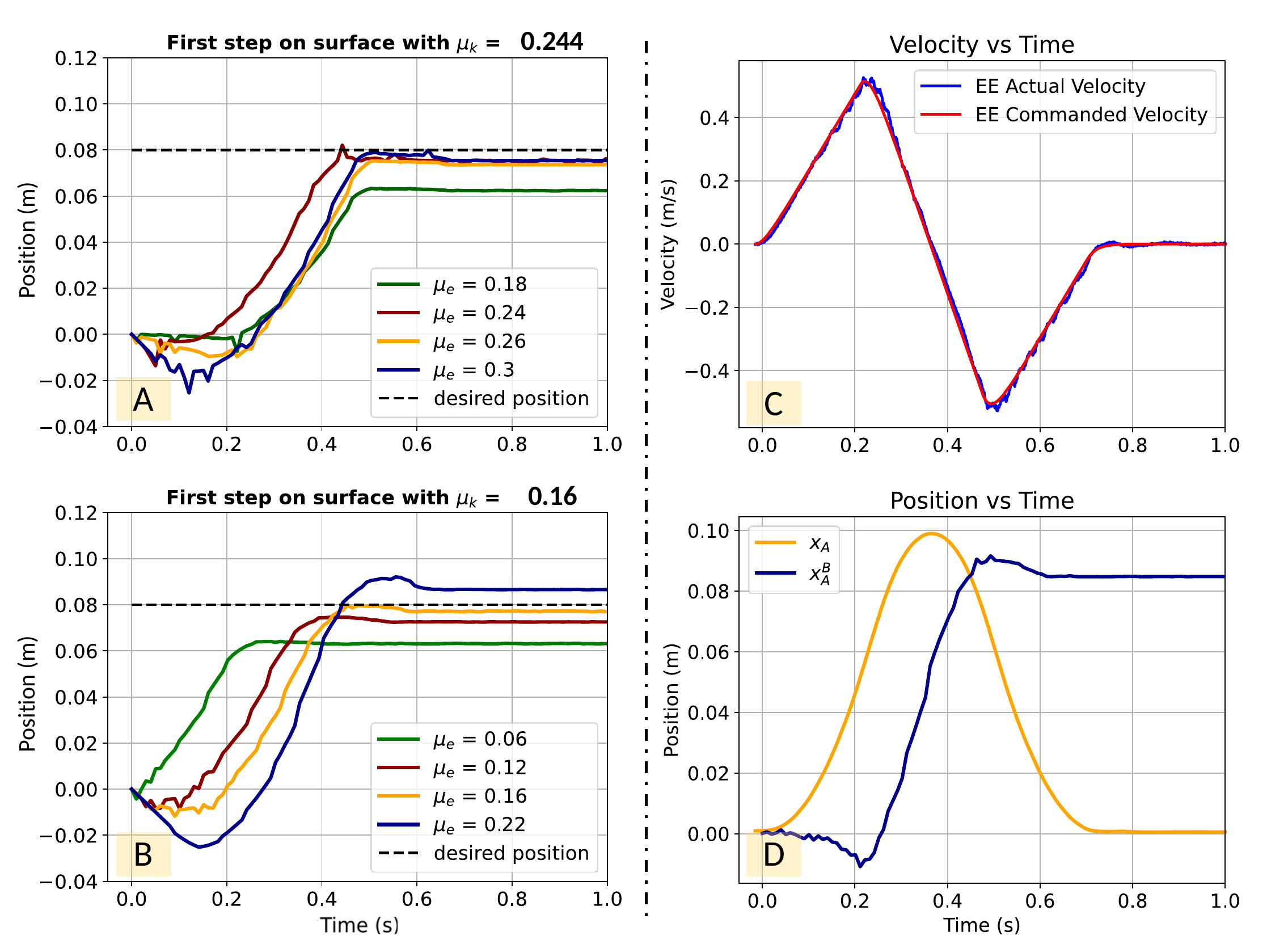}
    \caption{\textbf{A and B:} Show the actor network performance on two surfaces with different friction properties respectively \textit{$\mu_k = 0.244$} and \textit{$\mu_k = 0.16$} to slide an object for 8 cm when various \textit{$\mu_e$} are given as state to the network. \textbf{C:} illustrates the commanded and actual velocity of the end-effector (surface) during a single action and \textbf{D:} indicates the platform displacement and relative displacement of the object on the surface within the same action as C.}
    \label{fig:hardware_test}
\end{figure}

\subsection{ Zero-shot Sim-to-real Transfer Evaluation}

Initially, we test the trained model, referred to as the target actor network, within the DDPG RL framework (see \ref{fig:training_framework}) on a real-world setup designed for sliding an object on a horizontal surface. We conduct evaluations on two different surfaces, each characterized by dynamic friction coefficients of \textit{$\mu_k$ = 0.244} and  \textit{$\mu_k$ = 0.16}. To evaluate the model's adaptability and precision, we supply it with four different estimated friction coefficients (\textit{$\mu_e$}) to determine its capability to effectively perform the sliding task in a single step regardless of the accuracy of provided (\textit{$\mu_e$}). Results of this experiment is illustrated in Fig. \ref{fig:hardware_test}-A and Fig. \ref{fig:hardware_test}-B. With a desired relative displacement of 8 cm, the results show that an accurate estimate of \textit{$\mu_e$} keeps the displacement within the acceptable error range (\textit{$\left | \Delta x^B_A - D_{des} \right | \leq 1 cm$}). Furthermore, even a slight mismatch between \textit{$\mu_k$} and \textit{$\mu_e$} (i.e., \textit{$-0.06 \leq \mu_k - \mu_e \leq 0.06$}) still yields only a minor error—remaining below \textit{2 cm}—for the same \textit{$D_{des}$} in the first step.

To clearly illustrate how the actor network achieves the relative displacement in a single step on the real-world setup, Fig.\ref{fig:hardware_test}-C presents the commanded velocity—derived from the generated action and \eqref{eq:velocity_time}—along with the end-effector’s velocity tracking. Meanwhile, the displacement of the end-effector surface (\textit{$x_A$}) and the resulting relative displacement of the object (\textit{$x^B_A$}) during the one-step maneuver are depicted in Fig.\ref{fig:hardware_test}-D. As shown in Fig.~\ref{fig:hardware_test}-C and Fig. \ref{fig:hardware_test}-D, the predefined conditions for the maneuver, stated in \ref{sec:assumptions}, are satisfied.

\begin{figure}
\centering
    \includegraphics[trim=0.01cm 0.01cm 0.01cm 0.15cm, clip, width=0.48\textwidth, height=0.13\textheight]{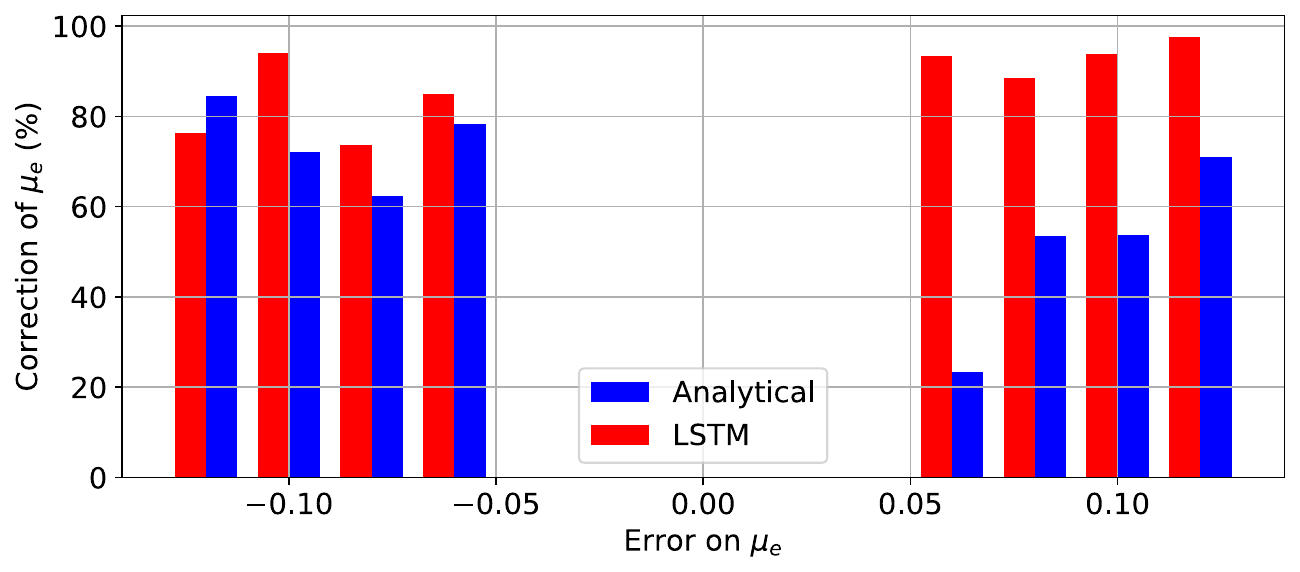}
    \caption{This plot illustrates how each friction estimation algorithm reduces the error in the estimated friction coefficient (\textit{$\mu_e$}), across a range of errors \textit{$|\mu_e - \mu_k|$} $\in  [0.06, 0.12] $ in real world experiment. The $y$ axis  shows the percentage improvement in the accuracy of the estimated friction coefficient for each model (LSTM and Analytical), where higher values indicate better performance.}
    \label{fig:LSTM_VS_ANALYTICAL_hard}
\end{figure}

Next, we seek to evaluate the accuracy of friction estimation algorithms on real world setup, where the required data stated in \ref{sec:algorithms}, particularly those regarding relative motion of the object are noisier than the simulation provided data. We extend the same experiment mentioned in \ref{sec:Sim_result} and reported for simulation setup in Fig. \ref{fig:friction_estimate_acc} to the real-world scenario. Fig. \ref{fig:LSTM_VS_ANALYTICAL_hard} compare the LSTM and the analytical approach performance in reducing the error between (\textit{$\mu_k$}) and (\textit{$\mu_e$}), when the DDPG trained actor network was supposed to perform an action for \textit{$D_{des} = 8\,cm$} in an environment characterized by \textit{$\mu_k = 0.16$}. Following this action, (\textit{$x_A^B$}), (\textit{$\dot{x}_A^B$}), and (\textit{$\ddot{x}_A$}) were provided to the algorithms. The results indicate, the LSTM approach has a better accuracy for the given conditions.  

\begin{figure}
\centering
    \includegraphics[trim=0.01cm 0.01cm 0.01cm 0.15cm, clip, width=0.48\textwidth, height=0.16\textheight]{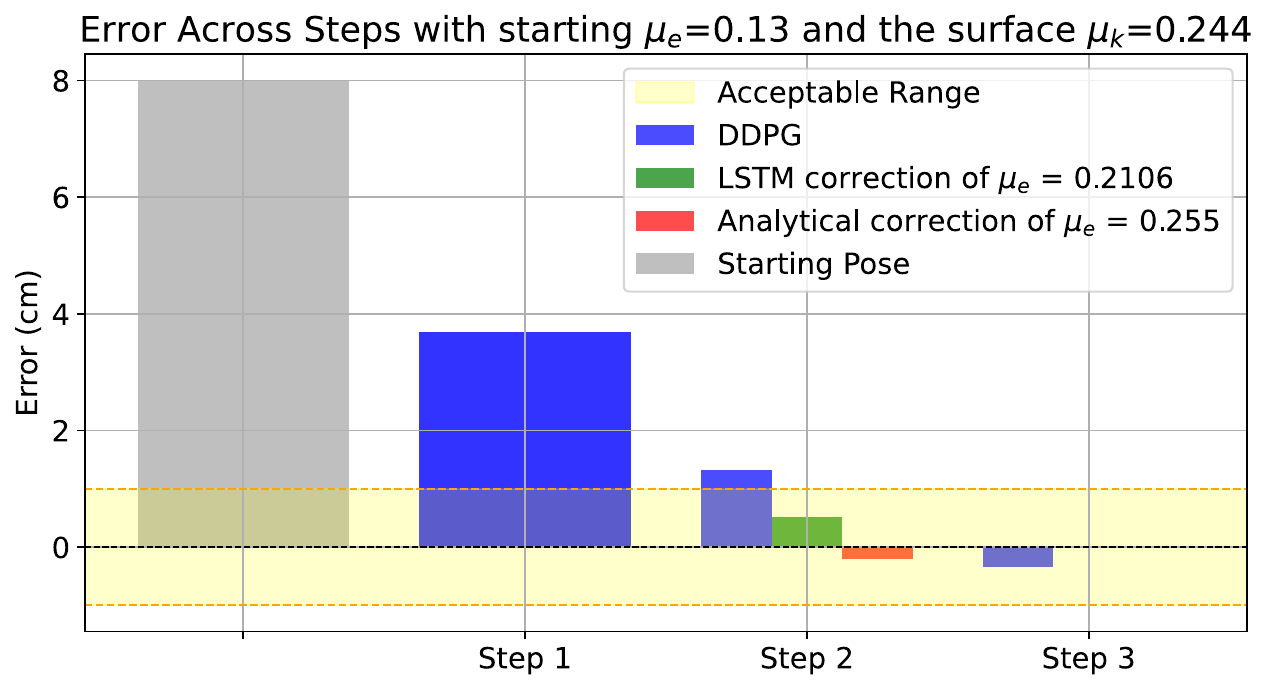}
    \caption{Comparison of the impact of friction estimation algorithms (analytical and LSTM) on task performance when the actor network starts with significant friction estimation error in simulation experiment.}
    \label{fig:DDPGvsLSTMvsANALY_Hard}
\end{figure}

Additionally, in the real-world setup, we examined a scenario where \textit{$\mu_k = 0.244$} and \textit{$\mu_e = 0.13$} was provided as the initial state to the actor network. As shown in the analogous simulation scenario in Fig. \ref{fig:sim_estimation_effect} the hardware tests yielded similar performance improvements. The friction estimation algorithms adjusted \textit{$\mu_e$} from 0.13 to \textit{$\mu_e = 0.2106$} using the LSTM approach and \textit{$\mu_e = 0.255$} using the analytical approach. These results demonstrate that the actor network's ability to compensate for discrepancies between \textit{$\mu_e$} and \textit{$\mu_k$}, thereby reducing the number of steps required to accomplish the sliding task and increasing the sliding accuracy in the subsequent step, extends beyond simulation and remains effective in real-world scenarios.

\section{Discussion and Conclusion}

The results of this study highlight the efficacy of the proposed DDPG RL framework in addressing the challenges of non-prehensile manipulation through sliding. By integrating friction estimation algorithms—an analytical approach and a data-driven LSTM model—the framework demonstrated robust performance in both simulation and real-world environments. The zero-shot sim-to-real transfer capability underscores the model's adaptability, and domain randomization effectiveness in bridging the gap between simulation and real world condition despite inaccuracies in the Coulomb friction model used in the simulation. Real-time friction estimation improved the accuracy of sliding displacements and reduced the need for corrective actions, as shown in Fig. \ref{fig:sim_estimation_effect} and Fig. \ref{fig:DDPGvsLSTMvsANALY_Hard}. Moreover, the LSTM friction estimation approach outperformed the analytical approach, emphasizing the value of data-driven techniques in analyzing dynamic and complex physical interactions. 

In conclusion, this study demonstrates the potential of DDPG RL framework, in non-prehensile manipulation through training in physics-accurate simulation environment and deployment in real world scenario. By paving the way for applications beyond sliding, this framework sets a foundation for advancing robotic dexterity in tasks traditionally dominated by human skills.

Future studies could focus on extending this framework to more complex and dynamic non-prehensile manipulation tasks such as flipping, rolling, or throwing. Improving robustness against higher levels of environmental noise and broadening the friction estimation algorithms to include additional factors—such as temperature or surface deformation—will further expand its practical applicability.



\balance





\end{document}